\documentclass[runningheads]{llncs}
\usepackage[T1]{fontenc}
\usepackage{graphicx}
\usepackage[utf8]{inputenc}
\usepackage{amsmath,amssymb}
\usepackage{esvect}
\usepackage{verbatim}
\usepackage[numbers]{natbib}
\usepackage[colorlinks=true,linkcolor=blue,citecolor=blue,urlcolor=blue]{hyperref}
\DeclareUnicodeCharacter{03BB}{$\lambda$}
\DeclareUnicodeCharacter{221A}{$\sqrt{}$}
\DeclareUnicodeCharacter{00B7}{$\cdot$}
\DeclareUnicodeCharacter{2212}{-}
\DeclareUnicodeCharacter{00B1}{$\pm$}
\begin{document}
\title{GraphDiffMed: Knowledge-Constrained Differential Attention with Pharmacological Graph Priors for Medication Recommendation}
\titlerunning{GraphDiffMed: Differential Attention with DDI Priors}
%
\author{Krati Saxena\inst{1}\orcidID{0000-0001-7049-9685} \and
Tomohiro Shibata\inst{1}\orcidID{0000-0002-8766-4250}}
\authorrunning{K. Saxena and T. Shibata}
%
\institute{
Kyushu Institute of Technology, Kitakyushu, Fukuoka, Japan\\
\email{saxena.krati536@mail.kyutech.jp, tom@brain.kyutech.ac.jp}}
\maketitle              
\begin{abstract}
Recommending safe and effective medication combinations from electronic health records (EHRs) is a core clinical AI problem, yet it remains difficult because patient trajectories are long, noisy, and clinically heterogeneous. Existing methods typically excel at either temporal modeling across visits or pharmacological knowledge integration (e.g., drug-drug interactions, DDIs), but rarely achieve both while robustly suppressing noise. We present GraphDiffMed, a knowledge-constrained medication recommendation framework built on dual-scale Differential Attention v2. Differential attention is applied at both intra-visit and inter-visit levels to filter spurious signals within encounters and across longitudinal history, while pharmacological constraints are incorporated during learning. Experiments on MIMIC-III and ablation studies show that this design consistently improves recommendation quality and ranking over strong baselines while achieving a more favorable safety performance balance. We further find that the strongest-performing configuration uses only demographic auxiliary features under our experimental setting. Overall, GraphDiffMed demonstrates that combining noise-aware attention with pharmacological constraints yields more reliable and clinically meaningful medication recommendation. We open-source our code at \url{https://github.com/saxenakrati09/GraphDiffMed}.

\end{abstract}
\section{Introduction}
\label{sect:introduction}
Medication recommendation is a central task in clinical decision support. Its goal is to recommend an appropriate set of medications for the current visit using longitudinal electronic health records (EHRs), including diagnoses, procedures, laboratory findings, and prior prescriptions, while controlling the risk of harmful drug-drug interactions (DDIs). The task is clinically important because modern care routinely involves polypharmacy, and clinicians must decide under time pressure, incomplete information, and evolving patient conditions. As a result, medication recommendation has become a standard benchmark in healthcare AI, with mature datasets, widely adopted evaluation protocols, and many competing methods.

Despite this maturity, the problem remains technically difficult. EHR data are sparse and noisy, with missing values and documentation inconsistencies that can create spurious patterns. Treatment effects unfold over time, so models must capture both local signals within a visit and long-range dependencies across visits. The output space is combinatorial because multiple medications are prescribed jointly, and safety cannot be separated from efficacy because of interaction risks. Meanwhile, substantial pharmacological knowledge exists in structured resources such as DDI graphs and molecular relations, but integrating it effectively with data-driven temporal modeling remains an open challenge.

Existing approaches provide partial solutions. Graph/knowledge-informed methods \citep{shang2019gamenet,liu2024safe,liang2025cidgmed} incorporate drug-interaction and/or molecular structure signals, but can still inherit biases from observational EHR data. Sequential and hierarchical EHR models \citep{choi2016retain,zhang2017leap,liu2023shape,wu2023dual} capture longitudinal visit dynamics (sometimes with patient similarity/attention), yet usually do not explicitly inject structured pharmacological knowledge. Hybrid/multimodal methods \citep{wu2024promise,bhoi2024refine,huo2024mifnet} fuse multiple information sources to improve recommendation quality, but their attention is often largely data-driven rather than pharmacologically constrained. In parallel, LLM-distillation approaches \citep{liu2024large} extract clinical semantics via prompting and distill them into smaller recommenders, providing an alternative route to external knowledge beyond explicit pharmacology graphs.

Conventional attention mechanisms often optimize predictive fit but do not reliably separate noisy co-occurrence patterns from clinically meaningful polypharmacy signals, potentially over-penalizing medication combinations that are clinically justified in complex cases. To address this quality-safety tension, we propose \textbf{GraphDiffMed}, a knowledge-constrained framework built on dual-scale Differential Attention v2 (DiffAttn\_v2). DiffAttn\_v2 is used at both intra-visit and inter-visit levels, and pharmacological constraints are incorporated as part of the full training framework.

Our contributions are as follows:
\begin{itemize}
\item We introduce the first dual-scale application of Differential Attention v2 for medication recommendation.
\item We analyze how knowledge constraints affect the safety-performance balance under different modality settings.
\item We provide a transparent decomposition of where gains originate through research question (RQ)-structured ablations.
\item We offer a clinical interpretation of the observed DDI profile, showing that slightly higher absolute DDI rates relative to conservative baselines can reflect more complete recommendations for complex polypharmacy cases.
\end{itemize}

The remainder of this paper is organized as follows. Section~\ref{sect:rel_works} reviews related work in medication recommendation, attention mechanisms, and knowledge integration. Section~\ref{sect:methodology} presents the GraphDiffMed architecture. Section~\ref{sect:experiments} describes datasets, settings, and evaluation metrics. Section~\ref{sect:results} reports main results and ablations. Section~\ref{sect:discussion} discusses clinical implications, limitations, and conclusions.

\section{Related Works}
\label{sect:rel_works}

\textbf{Medication recommendation models.} Deep learning methods have steadily improved temporal and structural modeling for medication recommendation. Early sequential models such as RETAIN \citep{choi2016retain} introduced interpretable RNN attention for clinical sequences, and LEAP \citep{zhang2017leap} cast prescribing as sequential decision making. To model complex interactions, graph-based methods including GAMENet \citep{shang2019gamenet} and SafeDrug \citep{liu2024safe} integrated external knowledge, explicitly representing drug-drug interaction (DDI) networks and molecular structure. Later architectures diversified: SHAPE \citep{liu2023shape}, DAPSNet \citep{wu2023dual}, and A-GSTCN \citep{yue2023gstcn} strengthened hierarchical temporal learning and patient similarity; CIDGMed \citep{liang2025cidgmed} used causal inference to correct historical bias; and multimodal models such as PROMISE \citep{wu2024promise} and MIFNet \citep{huo2024mifnet} fused heterogeneous EHR modalities. More recently, LEADER \citep{liu2024large} explored distilling clinical semantics from large language models. Despite strong empirical results, most methods still rely on largely data-driven temporal and structural attention, and even knowledge-aware models often weakly enforce validated pharmacological rules.

\textbf{Attention mechanisms and differential attention.} This limitation motivates closer examination of attention design itself. In healthcare modeling, standard paradigms-self-attention, cross-attention, and multi-head attention \citep{vaswani2017attention}-are widely used to emphasize relevant patient history. Yet these unconstrained mechanisms often overfit noise and institution-specific co-prescription artifacts in EHRs. To improve robustness, Differential Attention \citep{ye2024differential} introduced subtractive noise cancellation. Differential Transformer v2 \citep{ye2026difftransformerv2} extended this with query-dependent, per-token gating using sigmoid-constrained \(\lambda\) for fine-grained suppression. Although developed for natural language and vision, differential attention is now entering clinical modeling. Recently, DADA-MED \citep{saxena2025dada} added laboratory events and applied foundational differential attention at the intra-visit level. However, vanilla differential attention remains agnostic to external pharmacological structure, leaving it vulnerable to learning clinically unsafe correlations in rare drug combinations.

\textbf{Knowledge graph integration and DDI modeling.} This remaining gap points to explicit pharmacological grounding. Integrating external pharmacological knowledge is a standard approach to improving recommendation safety. Databases such as DrugBank \citep{knox2024drugbank} and TWOSIDES \citep{tatonetti2012data} are commonly used through embedding pretraining, graph neural network (GNN) message passing, or structural attention biasing. To reduce adverse events, state-of-the-art models (e.g., GAMENet \citep{shang2019gamenet}, PROMISE \citep{wu2024promise}, and REFINE \citep{bhoi2024refine}) usually frame DDI reduction as regularization by adding post hoc loss penalties. This penalty-centric design creates a clinical tension: trading broad DDI minimization against preserving therapeutically necessary, closely monitored polypharmacy. GraphDiffMed addresses this broader gap with a dual-scale differential-attention backbone and pharmacological constraints.

\section{Methodology}
\label{sect:methodology}

\subsection{Problem Formulation}

Let a patient's clinical history be represented as a sequence of visits: $\mathcal{P} = \{V_1, V_2, ..., V_{T-1}\}$, where $V_t$ denotes the $t$-th visit. Each visit $V_t$ contains:

\begin{itemize}
\item $D_t$: set of diagnoses (ICD codes)
\item $P_t$: set of procedures (CPT/ICD-9 procedure codes)
\item $M_t$: set of prescribed medications
\item $L_t$: set of laboratory events (test ID, value pairs)
\item $G_t$: patient gender (binary)
\item $A_t$: patient age at visit $t$
\end{itemize}

Given patient visits up to time $T$, the medication recommendation task predicts the medication set $M_T$ for visit $T$. Following standard benchmark practice in prior work, the model uses non-medication modalities from visits $1\ldots T$ (diagnoses/procedures in the default setting, and optionally demographics/laboratory events in additional-modality experiments), while medication inputs are restricted to visits $1\ldots T-1$. Thus, $M_T$ is never used as input, and no visits after $T$ are used.

\subsection{GraphDiffMed Architecture Overview}

\begin{figure}[!t]
\includegraphics[width=\textwidth]{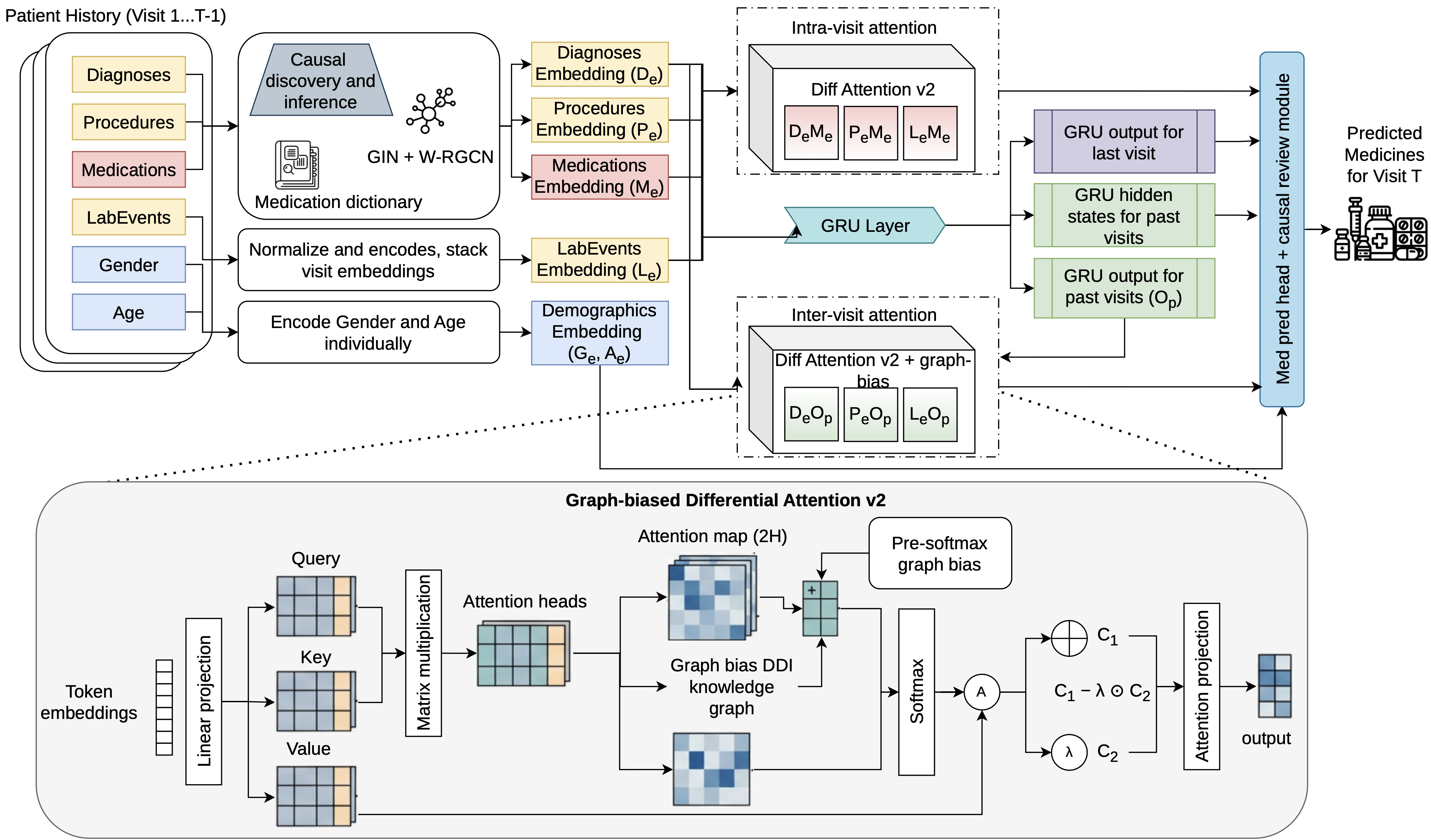}
\caption{Overview of GraphDiffMed} 
\label{fig:overview}
\end{figure}

GraphDiffMed follows a staged pipeline from representation learning to clinically informed prediction. First, a multi-modal embedding layer encodes diagnoses, procedures, medications, laboratory events, and demographic signals in a shared latent space. For intra-visit reasoning, each modality is represented as a single visit-level vector obtained by summing entity embeddings after graph processing, and graph-biased differential cross-attention computes a 1$\times$1 attention between the pooled medication vector (query) and pooled diagnosis/procedure vectors (key/value), effectively learning a gated projection of clinical context into medication representation space. Inter-visit encoding then captures longitudinal dependencies across historical visits. Effective graph bias is injected in inter-visit attention as a visit-set prior projected to medication-token positions. Finally, the aggregated patient representation is passed to a medication prediction head, followed by a causal review module that adjusts scores using diagnosis-medication and procedure-medication causal effects. An overview is shown in \autoref{fig:overview}.

\textbf{Multi-Modal Embedding Layer.}
This component follows prior designs and is adapted with explicit source separation. The modality processing pipeline for diagnoses, procedures, medications, laboratory events, and demographics is adopted from DADA-MED \citep{saxena2025dada}: diagnosis/procedure/medication codes are mapped to learnable embeddings ($\mathbf{D}_e,\mathbf{P}_e,\mathbf{M}_e$), laboratory events are encoded from normalized (test ID, value) pairs as $\mathbf{L}_e=\mathrm{ReLU}(\mathbf{W}_{\text{lab}}[\mathrm{ID}_{\text{norm}},\mathrm{value}_{\text{norm}}])$ and aggregated within each visit, and demographics are represented by a gender embedding and a linear age projection ($\mathbf{G}_e,\mathbf{A}_e$).

We use the CIDGMed pipeline \citep{liang2025cidgmed} for multimodal representation learning that jointly models diagnoses, procedures, and medications for visit-level recommendation (The second top box in \autoref{fig:overview} is adapted from CIDGMed). For medication representation, we draw on CIDGMed's fine-grained molecular branch, which maps each medication to molecular structures, performs GIN-style message passing over molecular nodes, and aggregates molecular information into medication embeddings through a learnable medication-molecule relation matrix. In CIDGMed, this medication branch is not used in isolation: diagnoses and procedures are also embedded separately and connected to medications through causal-effect matrices, after which dual-granularity representation learning integrates diagnosis, procedure, and medication information for visit-level recommendation.

\textbf{Graph-Biased Differential Attention.}
Our proposed method GraphDiffMed builds on DiffAttn\_v2 \citep{ye2026difftransformerv2} and injects pharmacological structure directly into attention formation. Let $\mathbf{X}$ be the query-side input and $\mathbf{Y}$ the key-value input. Following DiffAttn\_v2, queries are projected to doubled head space (for paired heads), i.e., $\mathbf{Q}=\mathbf{W}_Q\mathbf{X}$ with $2H$ heads, while keys/values are $\mathbf{K}=\mathbf{W}_K\mathbf{Y}$ and $\mathbf{V}=\mathbf{W}_V\mathbf{Y}$ with $H$ heads that are repeat-interleaved to align with $2H$ query heads. After reshaping, attention logits are
$$\mathbf{S}=\frac{\mathbf{Q}\mathbf{K}^T}{\sqrt{d_h}}+\lambda_{\text{graph}}\mathbf{B}_{\text{graph}}, \qquad
\mathbf{A}=\mathrm{softmax}(\mathbf{S}), \qquad
\mathbf{C}=\mathbf{A}\mathbf{V},$$
where $d_h$ is the per-head dimension, $\lambda_{\text{graph}}$ is a fixed scaling hyperparameter, and $\mathbf{B}_{\text{graph}}$ is the DDI-derived bias term. The context tensor $\mathbf{C}$ has $2H$ heads and is split into even/odd head pairs:
$$\mathbf{C}_1=\mathbf{C}[:,0::2,:,:], \qquad \mathbf{C}_2=\mathbf{C}[:,1::2,:,:].$$
Thus, $\mathbf{C}_1$ and $\mathbf{C}_2$ are not additional variables, but the two paired attention contexts produced by DiffAttn\_v2. A query-dependent gate is then computed per token and per head,
$$\boldsymbol{\lambda}=\sigma(\mathbf{W}_{\lambda}\mathbf{X}),$$
and differential denoising is applied as
$$\mathbf{C}_{\text{diff}}=\mathbf{C}_1-\boldsymbol{\lambda}\odot\mathbf{C}_2.$$
The final attention output is obtained with an output projection, $\mathbf{O}=\mathbf{W}_O\mathbf{C}_{\text{diff}}$. This subtraction suppresses shared noise while preserving query-relevant signal.

The key idea is that graph bias is added before softmax so pharmacological structure can shape attention before differential subtraction. In the current model, only the inter-visit bias matrix has this effect, because it varies across visit pairs. A single uniform scalar bias (as would be in the case of intra-visit attention) does not change attention weights after softmax, so graph bias is applied only in inter-visit attention. We use a visit-set-level bias projected to medication-related entries of the inter-visit attention matrix,
$$
\mathbf{B}_{\text{graph}}^{\text{inter}}[i,j]
=\mathbb{I}_{\text{med}}(i,j)
\cdot
\frac{1}{|M_{\text{query}}||M_{\text{key}}|}
\sum_{m_q\in M_{\text{query}}}
\sum_{m_k\in M_{\text{key}}}
\mathbf{A}_{DDI}[m_q,m_k].
$$
where $\mathbb{I}_{\text{med}}(i,j)=1$ if both positions $i$ and $j$ correspond to medication channels, and $0$ otherwise.
For each current-visit/historical-visit pair, we compute one scalar: the mean DDI density over all medication pairs across the two sets. This scalar is written to the corresponding entries of the inter-visit attention matrix. As a result, the prior can vary across historical visits, but not across individual drug pairs within a given visit pair. This is a design choice that follows the representation level: the temporal encoder attends over visit-pooled GRU states, not raw medication-token sequences. Therefore, the graph-prior effects observed in this study come from inter-visit bias at visit-set granularity. Achieving pair-level inter-visit bias would require attention over medication-level sequences instead of pooled visit states.

\textbf{Model Pipeline, Objective, and Implementation.}

The remaining modeling pipeline follows DADA-MED \citep{saxena2025dada}, while GraphDiffMed replaces only the attention operator with the graph-biased differential attention introduced above. Specifically, five components are retained: (1) \emph{intra-visit encoding}, where visit-level diagnosis, procedure, and prior-medication embeddings are aggregated as $\mathbf{h}_D^{(t)}=\sum_{d\in D_t}\mathbf{D}_e[d]$, $\mathbf{h}_P^{(t)}=\sum_{p\in P_t}\mathbf{P}_e[p]$, and $\mathbf{h}_M^{(t)}=\sum_{m\in M_{t-1}}\mathbf{M}_e[m]$, then refined by modality-specific homo-graphs (causal graphs for diagnoses/procedures and DDI graph for medications). Intra-visit differential cross-attention is then computed between pooled vectors (medication query, diagnosis/procedure key-value), yielding a 1$\times$1 attention map per head rather than token-sequence attention; (2) \emph{sequential temporal encoding} with modality-wise GRUs, e.g., $\mathbf{O}_D,\mathbf{h}_D^{\text{final}}=\mathrm{GRU}_D([\mathbf{h}_D^{(1)},\ldots,\mathbf{h}_D^{(T)}])$ (analogously for procedures and prior-medication states); (3) \emph{inter-visit attention}, where current-visit query $\mathbf{q}_{\text{visit}}$ (formed from current diagnosis/procedure states together with prior-medication and demographic embeddings $\mathbf{G}_e,\mathbf{A}_e$) attends over historical key-value context $\mathbf{kv}_{\text{prev}}$ (from $\mathbf{O}_D,\mathbf{O}_P,\mathbf{O}_M$) with $\mathbf{B}_{\text{graph}}^{\text{inter}}$; (4) \emph{patient representation aggregation}, where final hidden states, demographic embeddings, intra-/inter-visit attention summaries, and last-visit states are concatenated into $\mathbf{r}_{\text{patient}}$; and (5) \emph{medication prediction with causal review}, where logits $\mathbf{z}=\mathbf{W}_{\text{out}}\,\mathrm{ReLU}(\mathbf{r}_{\text{patient}})$ are adjusted using diagnosis/procedure-to-medication causal effects (via $c_m^{D/P}=\max_{d/p\in D_T/P_T}\mathrm{Causal Effect}(d/p,m)$) before obtaining probabilities $\mathbf{p}=\sigma(\mathbf{z})$. Key symbols used in this block are summarized in Table~\ref{tab:problem-notation}.

\begin{table}[t]
\centering
\caption{Notation used in model pipeline and objective.}
\label{tab:problem-notation}
\small
\setlength{\tabcolsep}{4pt}
\begin{tabular}{@{}p{0.3\textwidth}p{0.66\textwidth}@{}}
\hline
Symbol & Description \\
\hline
$D_t, P_t, M_t$ & diagnosis, procedure, and medication sets at visit $t$ \\
$\mathcal{V}_D,\mathcal{V}_P,\mathcal{V}_M$ & vocabularies of diagnoses, procedures, and medications \\
$vocabulary$ & complete set of unique coded tokens used as indices/labels \\
$n_D, n_P, n_M$ & corresponding vocabulary sizes \\
$\mathbf{D}_e,\mathbf{P}_e,\mathbf{M}_e,\mathbf{G}_e,\mathbf{A}_e$ & embedding tables/vectors used in the model (with code lookups such as $\mathbf{D}_e[d]$, $\mathbf{P}_e[p]$, and $\mathbf{M}_e[m]$) \\
$\mathbf{h}_\cdot^{(t)}$ & visit-level modality representations \\
$\mathbf{O}_\cdot, \mathbf{h}_\cdot^{\text{final}}$ & GRU output sequences and final hidden states \\
$\mathbf{q}_{\text{visit}}, \mathbf{kv}_{\text{prev}}$ & current-visit query and historical key-value tensors \\
$\mathbf{o}_{\text{intra}}, \mathbf{o}_{\text{inter}}$ & intra-visit and inter-visit attention outputs \\
$\mathbf{r}_{\text{patient}}$ & final concatenated patient representation for prediction \\
$\mathbf{A}_{DDI}\in\{0,1\}^{n_M\times n_M}$ & binary DDI adjacency matrix \\
$\mathbf{A}_{DDI}[i,j]=1$ & medications $i$ and $j$ have a known interaction \\
$d$ & embedding dimension \\
$y_m\in\{0,1\}, p_m$ & ground-truth label and predicted probability for medication $m$ \\
$M_{\text{pred}}$ & predicted medication set \\
$\theta$ & all trainable parameters \\
\hline
\end{tabular}
\normalsize
\end{table}

Training uses the multi-term objective,
$$\mathcal{L}=\mathcal{L}_{\text{BCE}}+\beta(t)\mathcal{L}_{\text{DDI}}+\alpha\mathcal{L}_{\text{reg}},$$
with binary cross-entropy
$$\mathcal{L}_{\text{BCE}}=-\frac{1}{n_M}\sum_{m=1}^{n_M}\left[y_m\log p_m+(1-y_m)\log(1-p_m)\right],$$
DDI regularization
$$\mathcal{L}_{\text{DDI}}=\frac{0.0005}{|M_{\text{pred}}|^2}\sum_{i,j\in M_{\text{pred}}}p_i p_j\mathbf{A}_{DDI}[i,j],$$
annealed DDI weighting
$$\beta(t)=\beta_0\left(1-\exp\left(-\gamma\frac{\mathrm{DDI}_{\text{current}}-\mathrm{DDI}_{\text{target}}}{\mathrm{DDI}_{\text{target}}}\right)\right),$$
and L2 regularization $\mathcal{L}_{\text{reg}}=\|\theta\|_2^2$, where $\mathrm{DDI}_{\text{target}}=0.06$ and $\gamma=2.5$. The annealed $\mathcal{L}_{\text{DDI}}$ term is active during training in all reported variants and directly optimizes lower interacting co-prescriptions at the output level. To avoid ambiguity, safety-aware behavior in GraphDiffMed comes from two complementary mechanisms: (i) output-level annealed DDI regularization (direct optimization), and (ii) attention-level graph bias (a pharmacological structural prior whose effect on DDI is emergent and configuration-dependent rather than directly optimized). Here, $t$ is the training step (or epoch) index; $\beta(t)$ is the dynamic weight for the DDI penalty; $\beta_0$ is the base DDI-penalty coefficient; and $\alpha$ is the L2-regularization coefficient. In $\mathcal{L}_{\text{BCE}}$, $n_M$ is the medication-vocabulary size, $m$ indexes medications, $y_m\in\{0,1\}$ is the ground-truth multi-label target for medication $m$, and $p_m\in(0,1)$ is its predicted probability. In $\mathcal{L}_{\text{DDI}}$, $M_{\text{pred}}$ is the predicted medication set, $|M_{\text{pred}}|$ is its cardinality, $i,j\in M_{\text{pred}}$ index predicted medications, $p_i,p_j$ are their predicted probabilities, and $\mathbf{A}_{DDI}[i,j]\in\{0,1\}$ indicates whether the pair has a known interaction. $\mathrm{DDI}_{\text{current}}$ is the current batch/model DDI rate, $\mathrm{DDI}_{\text{target}}$ is the desired target DDI rate, $\gamma$ controls annealing sharpness, and $\theta$ denotes all trainable parameters.

Implementation is configured with embedding dimension $d=64$, differential attention heads $H=8$ (thus $2H=16$), GRU hidden size $64$, dropout $0.7$, and fixed graph-bias scale $\lambda_{\text{graph}}=0.1$ (not optimized during training). Training uses Adam with learning rate $5\times10^{-4}$, patient-level batching (one patient per batch), $20$ epochs, and regularization weight $\alpha=0.005$. Diagnoses, procedures, and medications are always included as standard modalities; we then evaluate additional-modality settings \textbf{G} (gender only), \textbf{GY} (demographics: gender+age), \textbf{L} (lab events only), and \textbf{LGY} (labs+demographics), plus a full setting using all modalities jointly.

\section{Experimental Setup}
\label{sect:experiments}

\begin{table}[t]
\centering
\caption{Dataset preprocessing, statistics, and split configuration.}
\label{tab:data-summary}
\setlength{\tabcolsep}{3pt}
\resizebox{\columnwidth}{!}{%
\begin{tabular}{lll}
\hline
Category & Item & Value \\
\hline
Preprocessing & Single-visit patients & Filtered out \\
 & Medication code system & ATC level 3 \\
 & Diagnosis code system & ICD-9 \\
 & Procedure code system & ICD-9 procedure codes \\
 & Laboratory events & MIMIC-III LABEVENTS (test ID, value) \\
 & Demographics & Gender (binary), age (years) \\
Statistics & Total patients & 6,350 \\
 & Total visits & 15,032 \\
 & Average visits per patient & 2.37 \\
 & Diagnosis vocabulary size ($n_D$) & 1,958 \\
 & Procedure vocabulary size ($n_P$) & 1,426 \\
 & Medication vocabulary size ($n_M$) & 145 \\
 & DDI pairs & 1,318 \\
Split & Train & 5,080 patients \\
 & Validation & 635 patients \\
 & Test & 635 patients \\
\hline
\end{tabular}
}
\end{table}

We evaluate on MIMIC-III \citep{johnson2016mimic}. Preprocessing, statistics, and data splits are summarized in Table~\ref{tab:data-summary}. We compare GraphDiffMed with DADA-MED \citep{saxena2025dada}, CIDGMed \citep{liang2025cidgmed}, LEADER \citep{liu2024large}, LEAP \citep{zhang2017leap}, REFINE \citep{bhoi2024refine}, MIFNet \citep{huo2024mifnet}, SHAPE \citep{liu2023shape}, DAPSNet \citep{wu2023dual}, PROMISE \citep{wu2024promise}, and A-GSTCN \citep{yue2023gstcn}. We report Jaccard (primary), DDI Rate, F1, and PRAUC, plus Avg \#Meds as an auxiliary prescribing-intensity indicator. Avg \#Meds is the mean number of medications recommended per visit and is reported only as an \emph{auxiliary} contextual metric (not a primary model-selection target as is standard in state-of-the-art reporting). Results are mean $\pm$ standard deviation over five seeds (1, 3, 16, 18, 1234); model selection uses validation Jaccard, and test estimates use bootstrap resampling (10 iterations, 80\% test subset each).

\subsubsection*{4.4 Research Questions}

Our experiments are designed to answer the following research questions:
\begin{itemize}
\item RQ1: How does GraphDiffMed compare against state-of-the-art medication recommendation methods across all metrics?
\item RQ2: What is the contribution of the graph bias component, isolated per modality configuration, through ablation studies?
\item RQ3: How do different modality combinations (demographics, lab events) affect performance, both with and without graph bias?
\item RQ4: What is the isolated contribution of applying DiffAttn\_v2 at both temporal scales (without graph bias) compared to the full GraphDiffMed model?
\item RQ5: Does architectural innovation (dual-scale differential attention + graph bias) provide gains over simpler data-side engineering such as training set augmentation?
\end{itemize}

\section{Results and Analysis}
\label{sect:results}
\subsubsection*{5.1 Main Results (RQ1)}

Table~\ref{tab:main-results} reports the primary comparison on MIMIC-III. GraphDiffMed (GY) is best on Jaccard, F1, and PRAUC, while PROMISE achieves the lowest DDI. This pattern indicates that the proposed model improves recommendation quality and ranking, with a safety profile that must be interpreted jointly with coverage rather than as an isolated minimum-DDI objective.

\begin{table*}[t]
\centering
\caption{Main results on MIMIC-III test set. The best performing model is \textbf{GraphDiffMed (GY)}.}
\label{tab:main-results}
\small
\setlength{\tabcolsep}{3pt}
\renewcommand{\arraystretch}{1.05}
\resizebox{\textwidth}{!}{%
\begin{tabular}{lccccc}
\hline
Method & Jaccard $\uparrow$ & DDI $\downarrow$ & F1 $\uparrow$ & PRAUC $\uparrow$ & Avg meds \\
\hline
CIDGMed & 0.5526$\pm$0.0016 & 0.0684$\pm$0.0014 & 0.7033$\pm$0.0009 & 0.7955$\pm$0.0026 & 22.47$\pm$0.00 \\
LEADER & 0.5391$\pm$0.0015 & - & 0.6921$\pm$0.0014 & 0.7816$\pm$0.0015 & - \\
LEAP & 0.4526$\pm$0.0007 & 0.0762$\pm$0.0015 & 0.6147$\pm$0.0021 & 0.6555$\pm$0.0014 & 18.62$\pm$0.00 \\
REFINE & 0.5235$\pm$0.0018 & - & 0.6794$\pm$0.0017 & 0.7791$\pm$0.0017 & - \\
MIFNet & 0.4461$\pm$0.0016 & 0.0810$\pm$0.0008 & 0.6031$\pm$0.0015 & 0.6859$\pm$0.0018 & 13.53$\pm$0.04 \\
SHAPE & 0.5513$\pm$0.0009 & 0.0677$\pm$0.0003 & 0.7017$\pm$0.0008 & 0.7906$\pm$0.0009 & 20.99$\pm$0.12 \\
DAPSNet & 0.5469$\pm$0.0025 & 0.0657$\pm$0.0010 & 0.6989$\pm$0.0013 & 0.7942$\pm$0.0024 & 21.07$\pm$0.21 \\
PROMISE & 0.5326$\pm$0.0018 & \textbf{0.0653$\pm$0.0006} & 0.6848$\pm$0.0021 & 0.7764$\pm$0.0020 & 19.76$\pm$0.09 \\
A-GSTCN & 0.4689 & - & 0.6307 & 0.7113 & 15.34 \\
\textbf{GraphDiffMed} & \textbf{0.5577$\pm$0.0010} & 0.0718$\pm$0.0010 & \textbf{0.7074$\pm$0.0009} & \textbf{0.8005$\pm$0.0007} & 22.10$\pm$0.18 \\
\hline
\end{tabular}
}
\end{table*}

\subsubsection*{5.2 Ablation and Modality Effects (RQ2, RQ3, RQ4)}

Table~\ref{tab:ablation-results} jointly answers RQ2, RQ3, and RQ4. Here, \emph{Dual v2} denotes the dual-scale DiffAttn\_v2 architecture (intra-visit + inter-visit differential attention) \emph{without} graph bias; GraphDiffMed adds graph bias on top of the same backbone. First, compared with the baseline (v1), Dual v2 variants improve Jaccard and F1, confirming that dual-scale denoising is the primary performance driver. Second, the baseline yields the lowest DDI, but this comes with clearly lower Jaccard and F1, indicating a conservative quality-safety point. Third, the best Jaccard/F1 are achieved by Dual v2 (LGY), but with higher DDI. Notably, GraphDiffMed (GY) attains the second-best Jaccard/F1 together with the best PRAUC and a substantially lower DDI than Dual v2 (LGY), indicating a strong quality-safety trade-off.

\begin{table*}[t]
\centering
\caption{Ablation across dual-scale DiffAttn\_v2 and graph bias. Here, \((- )\) denotes the default modality set \((D,P,M)\), while L, GY, and LGY indicate additional modalities appended to this base. Bold and italic values show the first and the second best results respectively.}
\label{tab:ablation-results}
\small
\setlength{\tabcolsep}{2.5pt}
\renewcommand{\arraystretch}{1.05}
\resizebox{\textwidth}{!}{%
\begin{tabular}{lccccc}
\hline
Variant & Jaccard $\uparrow$ & DDI $\downarrow$ & F1 $\uparrow$ & PRAUC $\uparrow$ & Avg meds \\
\hline
Baseline (v1) & 0.5549$\pm$0.0015 & \textbf{0.0711$\pm$0.0006} & 0.7049$\pm$0.0012 & 0.7976$\pm$0.0018 & 22.24$\pm$0.28 \\
Dual v2 (-) & 0.5573$\pm$0.0005 & 0.0717$\pm$0.0010 & 0.7070$\pm$0.0004 & 0.7995$\pm$0.0008 & 21.97$\pm$0.26 \\
Dual v2 (L) & 0.5560$\pm$0.0011 & 0.0716$\pm$0.0020 & 0.7059$\pm$0.0010 & 0.7972$\pm$0.0007 & 21.47$\pm$0.53 \\
Dual v2 (GY) & 0.5577$\pm$0.0012 & 0.0724$\pm$0.0024 & 0.7073$\pm$0.0010 & \textit{0.7998$\pm$0.0023} & 22.08$\pm$0.33 \\
Dual v2 (LGY) & \textbf{0.5579$\pm$0.0007} & 0.0724$\pm$0.0010 & \textbf{0.7075$\pm$0.0005} & 0.7991$\pm$0.0009 & 21.88$\pm$0.29 \\
GraphDiffMed (-) & 0.5569$\pm$0.0009 & \textit{0.0716$\pm$0.0017} & 0.7065$\pm$0.0008 & 0.7994$\pm$0.0010 & 22.18$\pm$0.25 \\
GraphDiffMed (L) & 0.5571$\pm$0.0018 & 0.0717$\pm$0.0013 & 0.7068$\pm$0.0016 & 0.7973$\pm$0.0016 & 21.72$\pm$0.17 \\
\textbf{GraphDiffMed (GY)} & \textit{0.5577$\pm$0.0010} & 0.0718$\pm$0.0010 & \textit{0.7074$\pm$0.0009} & \textbf{0.8005$\pm$0.0007} & 22.10$\pm$0.18 \\
GraphDiffMed (LGY) & 0.5576$\pm$0.0004 & 0.0724$\pm$0.0013 & 0.7072$\pm$0.0003 & 0.7986$\pm$0.0014 & 21.75$\pm$0.26 \\
\hline
\end{tabular}
}
\end{table*}

\subsubsection*{5.3 Comparison with Data Augmentation (RQ5)}

Table~\ref{tab:aug-results} compares the augmentation-only variants against the proposed architecture. Relative to Table~\ref{tab:main-results} and Table~\ref{tab:ablation-results}, augmentation alone yields smaller gains, indicating that architectural changes (dual-scale differential attention with graph bias) are the dominant source of improvement.

\begin{table}[t]
\centering
\caption{Augmentation-only variants (without graph-biased differential attention).}
\label{tab:aug-results}
\setlength{\tabcolsep}{4pt}
\resizebox{\columnwidth}{!}{%
\begin{tabular}{lcccc}
\hline
Method & Jaccard $\uparrow$ & DDI $\downarrow$ & F1 $\uparrow$ & PRAUC $\uparrow$ \\
\hline
Aug-Med (GY) & 0.5552$\pm$0.0009 & 0.0717$\pm$0.0005 & 0.7052$\pm$0.0007 & 0.7981$\pm$0.0015 \\
Aug-Med (L) & 0.5560$\pm$0.0011 & 0.0711$\pm$0.0015 & 0.7058$\pm$0.0010 & 0.7978$\pm$0.0015 \\
Aug-Med (LGY) & 0.5558$\pm$0.0010 & 0.0722$\pm$0.0007 & 0.7057$\pm$0.0010 & 0.7982$\pm$0.0012 \\
\hline
\end{tabular}
}
\end{table}

\subsubsection*{5.4 Statistical testing, Attention, Error Profile}
Statistical testing supports the same overall pattern. Augmentation-only variants are mostly non-significant on primary quality metrics (Jaccard, F1, PRAUC), with isolated effects mainly on DDI or Avg meds (e.g., LGY augmentation: DDI \(p=0.0086\)). In contrast, dual-scale DiffAttn\_v2 variants show consistent and significant gains in Jaccard and F1 versus baseline for the base setting \((D,P,M)\), GY, and LGY, both without and with graph bias. The strongest significance appears in the graph-biased GY setting (Jaccard \(p=0.0030\), F1 \(p=0.0016\)). PRAUC significance is mainly observed in the base \((D,P,M)\) and GY settings, whereas DDI differences are usually non-significant except in LGY variants. Overall, the most reliable effect is improvement in predictive quality, while DDI trade-offs are concentrated in higher-modality settings.

Qualitatively, inter-visit attention is clinically coherent, preserving continuity of chronic high-risk medications; with graph bias, medication-to-medication focus is less diffuse and better aligned with known DDI-relevant pairs, supporting its role as a structural prior rather than a hard rule. Errors are also clinically plausible: false negatives are more common for rare or episodic medications, while false positives are often reasonable alternatives under physician-level prescribing variability. Some predicted interacting pairs may reflect overfitting, but many are combinations used in monitored ICU practice, consistent with the observed quality-safety trade-off.

\section{Discussion, Limitations, and Conclusion}
\label{sect:discussion}

GraphDiffMed improves medication recommendation primarily through dual-scale Differential Attention v2, which strengthens noise suppression within encounters and across longitudinal history. Ablations indicate that most Jaccard/F1 gains come from this architectural change, while knowledge constraints provide a secondary, configuration-dependent contribution to the quality-safety balance (most clearly in the GY setting).

The empirical profile is clinically meaningful rather than purely metric-driven. Within ablations, Dual v2 (LGY) reaches the top Jaccard and F1, but at a higher DDI level. GraphDiffMed (GY), which uses only demographics as auxiliary inputs, preserves near-peak Jaccard/F1 (second-best), achieves the best PRAUC, and lowers DDI relative to Dual v2 (LGY). The baseline remains the lowest-DDI setting, but its lower Jaccard/F1 indicates under-recommendation. In ICU practice, some interacting pairs are clinically necessary and managed through monitoring, dose adjustment, and scheduling; accordingly, the binary DDI metric should be interpreted together with recommendation completeness because it does not distinguish contraindicated from manageable interactions.

This demographics finding is notable. Age and gender provide stable, low-noise stratification signals that are broadly available across EHR systems, whereas laboratory features are high-dimensional, sparse, and temporally noisy in MIMIC-III ICU trajectories. Under fixed model capacity and strong regularization, cleaner auxiliary signals appear to be more useful than noisier high-dimensional inputs. This also helps explain why graph bias contributes more reliably in cleaner-feature configurations and is diluted when noisy modalities dominate attention patterns.

Several limitations remain. Evaluation is restricted to MIMIC-III ICU data, so external validity to MIMIC-IV, eICU, and outpatient settings is still open. ATC level-3 aggregation simplifies medication space and omits dosage/ formulation granularity. Agreement-based metrics (Jaccard/F1/PRAUC) proxy clinician prescribing behavior rather than causal treatment optimality, and prospective outcome validation is still needed. The DDI knowledge used in this study follows the baseline preprocessing pipeline (e.g., CIDGMed) and is a binary adjacency matrix without severity or mechanism weights; therefore, severity-aware comparison would require matched reimplementation across baselines for strict fairness. In addition, inter-visit graph bias is computed at visit-set granularity on pooled visit states, not individual drug-pair granularity, which limits influence relative to explicit pairwise DDI constraints. We also fix $\lambda_{\text{graph}}$ as a hyperparameter ($0.1$) rather than learning it, which may understate or overstate the attainable impact of the graph prior. 

Future work should prioritize adding an explicit intra-visit medication-pair graph-bias mechanism (e.g., a sparse current-visit DDI submatrix over medication tokens), implementing medication-level (non-pooled) inter-visit attention to enable pair-level cross-visit graph bias, severity-aware DDI modeling, multitask learning with adverse-event prediction, personalized DDI risk, cross-dataset validation, stronger explanation methods (including counterfactuals), and prospective longitudinal evaluation; integration with language models for clinician-facing explanations and online adaptation are additional promising directions.

In summary, GraphDiffMed demonstrates that dual-scale differential denoising is the main driver of improved recommendation quality, with knowledge constraints contributing to safety-performance trade-offs in selected settings. These trends are supported by statistical testing: Jaccard/F1 gains are consistently significant for Dual v2 and GraphDiffMed in the GY and LGY settings, whereas DDI differences are mostly non-significant except in higher-modality LGY variants. The results show that clean, interpretable auxiliary features and noise-robust attention can outperform more complex but noisier multimodal settings.

\bibliographystyle{splncs04}
\bibliography{citations}

@article{huo2024mifnet,
  title={MIFNet: multimodal interactive fusion network for medication recommendation},
  author={Huo, Jiazhen and Hong, Zhikai and Chen, Mingzhou and Duan, Yongrui},
  journal={The Journal of Supercomputing},
  pages={1--33},
  year={2024},
  publisher={Springer}
}

@article{liang2025cidgmed,
  title={CIDGMed: Causal Inference-Driven Medication Recommendation with Enhanced Dual-Granularity Learning},
  author={Liang, Shunpan and Li, Xiang and Mu, Shi and Li, Chen and Lei, Yu and Hou, Yulei and Ma, Tengfei},
  journal={Knowledge-Based Systems},
  volume={309},
  pages={112685},
  year={2025},
  publisher={Elsevier}
}

@article{liu2024large,
  title={Large language model distilling medication recommendation model},
  author={Liu, Qidong and Wu, Xian and Zhao, Xiangyu and Zhu, Yuanshao and Zhang, Zijian and Tian, Feng and Zheng, Yefeng},
  journal={arXiv preprint arXiv:2402.02803},
  year={2024}
}

@article{liu2024safe,
  title={Safe drug recommendation through forward data imputation and recurrent residual neural network},
  author={Liu, Junping and Wan, Zhiju and Hu, Xinrong and Zhu, Qiang},
  journal={Applied Soft Computing},
  volume={161},
  pages={111723},
  year={2024},
  publisher={Elsevier}
}

@article{wu2024promise,
  title={PROMISE: A pre-trained knowledge-infused multimodal representation learning framework for medication recommendation},
  author={Wu, Jialun and Yu, Xinyao and He, Kai and Gao, Zeyu and Gong, Tieliang},
  journal={Information Processing \& Management},
  volume={61},
  number={4},
  pages={103758},
  year={2024},
  publisher={Elsevier}
}

@article{liu2023shape,
  title={SHAPE: A Sample-adaptive Hierarchical Prediction Network for Medication Recommendation},
  author={Liu, Sicen and Wang, Xiaolong and Du, Jingcheng and Hou, Yongshuai and Zhao, Xianbing and Xu, Hui and Wang, Hui and Xiang, Yang and Tang, Buzhou},
  journal={IEEE Journal of Biomedical and Health Informatics},
  year={2023},
  publisher={IEEE}
}

@article{wu2023dual,
  title={Dual attention and patient similarity network for drug recommendation},
  author={Wu, Jialun and Dong, Yuxin and Gao, Zeyu and Gong, Tieliang and Li, Chen},
  journal={Bioinformatics},
  volume={39},
  number={1},
  pages={btad003},
  year={2023},
  publisher={Oxford University Press}
}

@article{yue2023gstcn,
  title={A-GSTCN: An Augmented Graph Structural--Temporal Convolution Network for Medication Recommendation Based on Electronic Health Records},
  author={Yue, Weiqi and Wang, Maiqiu and Zhang, Lei and Zhang, Lijuan and Huang, Jie and Wan, Jian and Xiong, Naixue and Vasilakos, Athanasios V},
  journal={Bioengineering},
  volume={10},
  number={11},
  pages={1241},
  year={2023},
  publisher={MDPI}
}

@inproceedings{zhang2017leap,
  title={LEAP: learning to prescribe effective and safe treatment combinations for multimorbidity},
  author={Zhang, Yutao and Chen, Robert and Tang, Jie and Stewart, Walter F and Sun, Jimeng},
  booktitle={proceedings of the 23rd ACM SIGKDD international conference on knowledge Discovery and data Mining},
  pages={1315--1324},
  year={2017}
}

@article{bhoi2024refine,
  title={REFINE: a fine-grained medication recommendation system using deep learning and personalized drug interaction modeling},
  author={Bhoi, Suman and Lee, Mong Li and Hsu, Wynne and Tan, Ngiap Chuan},
  journal={Advances in Neural Information Processing Systems},
  volume={36},
  year={2024}
}

@online{ye2026difftransformerv2,
  title = {Differential Transformer V2},
  author = {Ye, Tianzhu and Dong, Li and Sun, Yutao and Wei, Furu},
  year = {2026},
  month = {1},
  day = {20},
  url = {https://aka.ms/diff-transformer-v2},
  urldate = {2026-1-20}
}

@article{vaswani2017attention,
  title={Attention is all you need},
  author={Vaswani, A},
  journal={Advances in Neural Information Processing Systems},
  year={2017}
}

@article{johnson2016mimic,
  title={MIMIC-III, a freely accessible critical care database},
  author={Johnson, Alistair EW and Pollard, Tom J and Shen, Lu and Lehman, Li-wei H and Feng, Mengling and Ghassemi, Mohammad and Moody, Benjamin and Szolovits, Peter and Anthony Celi, Leo and Mark, Roger G},
  journal={Scientific data},
  volume={3},
  number={1},
  pages={1--9},
  year={2016},
  publisher={Nature Publishing Group}
}

@inproceedings{saxena2025dada,
  title={DADA-MED: Data-Augmented Dual Attention Model for Enhanced Medication Recommendations},
  author={Saxena, Krati and Shibata, Tomohiro},
  booktitle={IFIP International Conference on Artificial Intelligence Applications and Innovations},
  pages={83--97},
  year={2025},
  organization={Springer}
}

@inproceedings{shang2019gamenet,
  title={Gamenet: Graph augmented memory networks for recommending medication combination},
  author={Shang, Junyuan and Xiao, Cao and Ma, Tengfei and Li, Hongyan and Sun, Jimeng},
  booktitle={proceedings of the AAAI Conference on Artificial Intelligence},
  volume={33},
  pages={1126--1133},
  year={2019}
}

@article{choi2016retain,
  title={Retain: An interpretable predictive model for healthcare using reverse time attention mechanism},
  author={Choi, Edward and Bahadori, Mohammad Taha and Sun, Jimeng and Kulas, Joshua and Schuetz, Andy and Stewart, Walter},
  journal={Advances in neural information processing systems},
  volume={29},
  year={2016}
}

@article{ye2024differential,
  title={Differential transformer},
  author={Ye, Tianzhu and Dong, Li and Xia, Yuqing and Sun, Yutao and Zhu, Yi and Huang, Gao and Wei, Furu},
  journal={arXiv preprint arXiv:2410.05258},
  year={2024}
}

@article{knox2024drugbank,
  title={DrugBank 6.0: the DrugBank knowledgebase for 2024},
  author={Knox, Craig and Wilson, Mike and Klinger, Christen M and Franklin, Mark and Oler, Eponine and Wilson, Alex and Pon, Allison and Cox, Jordan and Chin, Na Eun and Strawbridge, Seth A and others},
  journal={Nucleic acids research},
  volume={52},
  number={D1},
  pages={D1265--D1275},
  year={2024},
  publisher={Oxford University Press}
}

@article{tatonetti2012data,
  title={Data-driven prediction of drug effects and interactions},
  author={Tatonetti, Nicholas P and Ye, Patrick P and Daneshjou, Roxana and Altman, Russ B},
  journal={Science translational medicine},
  volume={4},
  number={125},
  pages={125ra31--125ra31},
  year={2012},
  publisher={American Association for the Advancement of Science}
}
%




\end{document}